\documentclass[conference]{IEEEtran}
\IEEEoverridecommandlockouts
\usepackage{cite}
\usepackage{amsmath,amssymb,amsfonts}
\usepackage{algorithmic}
\usepackage{graphicx}
\usepackage{textcomp}
\usepackage{xcolor}
\usepackage{graphicx}
\usepackage{amsmath}
\usepackage{diagbox}
\usepackage{stfloats}
\usepackage{algorithm}
\usepackage{multirow}
\usepackage[flushleft]{threeparttable}

\def\BibTeX{{\rm B\kern-.05em{\sc i\kern-.025em b}\kern-.08em
    T\kern-.1667em\lower.7ex\hbox{E}\kern-.125emX}}
\begin{document}

\title{Pseudo-Labeling for Small Lesion Detection on Diabetic Retinopathy Images\\
}

\makeatletter
\newcommand{\linebreakand}{%
  \end{@IEEEauthorhalign}
  \hfill\mbox{}\par
  \mbox{}\hfill\begin{@IEEEauthorhalign}
}
\makeatother
\author{\IEEEauthorblockN{Qilei Chen}
\IEEEauthorblockA{\textit{Department of Computer Science} \\
\textit{University of Massachusetts Lowell}\\
Lowell, MA, USA \\
qilei\_chen@student.uml.edu}
\and
\IEEEauthorblockN{Ping Liu}
\IEEEauthorblockA{\textit{Department of Ophthalmology} \\
\textit{Central South University}\\
Changsha, Hunan, China \\
liupinglucy@csu.edu.cn}
\and
\IEEEauthorblockN{Jing Ni}
\IEEEauthorblockA{\textit{Department of Computer Science} \\
\textit{University of Massachusetts Lowell}\\
Lowell, MA, USA \\
jing\_ni@student.uml.edu}
\linebreakand
\IEEEauthorblockN{Yu Cao}
\IEEEauthorblockA{\textit{Department of Computer Science} \\
\textit{University of Massachusetts Lowell}\\
Lowell, MA, USA \\
ycao@cs.uml.edu}
\and
\IEEEauthorblockN{Benyuan Liu}
\IEEEauthorblockA{\textit{Department of Computer Science} \\
\textit{University of Massachusetts Lowell}\\
Lowell, MA, USA \\
bliu@cs.uml.edu}
\and
\IEEEauthorblockN{Honggang Zhang}
\IEEEauthorblockA{\textit{Department of Engineering} \\
\textit{University of Massachusetts Boston}\\
Boston, MA, USA \\
Honggang.Zhang@umb.edu}
}

\maketitle


\begin{abstract}
Diabetic retinopathy (DR) is a primary cause of blindness in working-age people worldwide. About 3 to 4 million people with diabetes become blind because of DR every year. Diagnosis of DR through color fundus images is a common approach to mitigate such problem. However, DR diagnosis is a difficult and time consuming task, which requires experienced clinicians to identify the presence and significance of many small features on high resolution images. Convolutional Neural Network (CNN) has proved to be a promising approach for automatic biomedical image analysis recently. In this work, we investigate lesion detection on DR fundus images with CNN-based object detection methods. Lesion detection on fundus images faces two unique challenges. The first one is that our dataset is not fully labeled, i.e., only a subset of all lesion instances are marked. Not only will these unlabeled lesion instances not contribute to the training of the model, but also they will be mistakenly counted as false negatives, leading the model move to the opposite direction. The second challenge is that the lesion instances are usually very small, making them difficult to be found by normal object detectors. To address the first challenge, we introduce an iterative training algorithm for the semi-supervised method of pseudo-labeling, in which a considerable number of unlabeled lesion instances can be discovered to boost the performance of the lesion detector. For the small size targets problem, we extend both the input size and the depth of feature pyramid network (FPN) to produce a large CNN feature map, which can preserve the detail of small lesions and thus enhance the effectiveness of the lesion detector. The experimental results show that our proposed methods significantly outperform the baselines.

\end{abstract}


\section{Introduction}
Diabetic Retinopathy (DR) is becoming a leading cause of ophthalmic diseases globally 
and it is one of the most common complications of diabetes. DR is caused by diabetes as a result of retinal blood vessel damage by elevated glucose levels. 
The worst situation can be total blindness. It is reported that the prevalence of DR in diabetic population has reached 37.5\% and an estimated 370 million people worldwide will be affected by diabetes mellitus by 2030 \cite{danaei2011national}.
Research has found that early treatment is an effective approach to reduce the risk of blindness \cite{zhang2017prevalence}. During treatment, early screening and regular checkups with fundus camera are essential steps but these are difficult and time-consuming tasks for clinicians. Therefore, 
it is very important to develop an effective tool for DR diagnosis in early screening to improve the healthcare outcome. 

International Clinical Diabetic Retinopathy and Diabetic Macular Edema Disease Severity Scales (ICDRDMEDSS) is a worldwide-used standard for severity diagnosis based on DR images \cite{wilkinson2003proposed}, in which the severity stage is determined by the location and number of different lesion instances on a fundus image. 
\begin{figure}
\centering
\includegraphics[width=0.8\columnwidth,height=\textwidth,keepaspectratio]{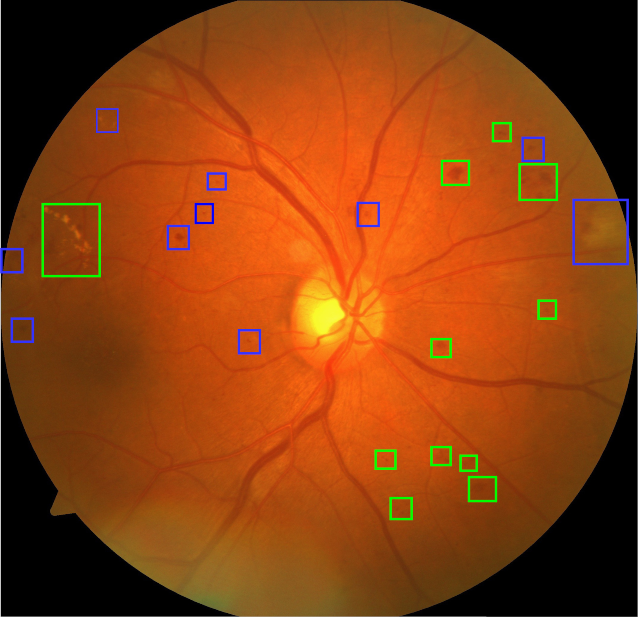}\caption{Green boxes represent the labeled lesion instances while blue boxes represent the lesion instances without labels.}\label{Fig2}
\end{figure}
With the success of images analysis methods based on Convolutional Neural Network (CNN) \cite{ren2015faster,redmon2016you,liu2016ssd,alcantara2017improving,sun2019people,girshick2015fast}, it is natural to consider using CNN models to automatically analyze fundus images. In fact, the diagnosis process can be considered as an instance level multi-label visual object detection task to find lesions and their corresponding categories in DR images, which has become a hot research topic in recent years \cite{dai2017retinal,wang2017zoom,yang2017lesion,zhao2018uniqueness}. 
However, most of the previous methods are designed to detect lesions of a single category (e.g., \cite{dai2017retinal,yang2017lesion}), or to detect multiple lesion categories in a sequence of steps \cite{zhao2018uniqueness}.  Due to the lack of datasets with detailed lesion location information, previous weakly supervised methods \cite{wang2017zoom} can only find suspicious lesion regions rather than fine grained individual lesion instances. 
We aim to develop a method based on CNN to detect all lesion instances of different categories with a single model in one round. 

We consider five publicly available retinal image datasets: Diabetic Retinopathy Database and Evaluation Protocol Version 2.1 (DRD\_EPV2.1) \cite{antal2012ensemble}; the e-ophtha \cite{decenciere2013teleophta}; Indian Diabetic Retinopathy Image Dataset (IDRID) \cite{porwal2018indian}; the Retina Check project managed by Eindhoven University of Technology (RC-RGB-MA)\cite{dashtbozorg2018retinal}; the Retinopathy Online Challenge training set (ROC)\cite{niemeijer2009retinopathy}. Due to the limited number of images in these datasets, they are not suitable for training CNN models. To this end, our team collected a large dataset containing about 5,000 DR screening color fundus images. 
It is a labor-intensive task to find and label every lesion instance of the dataset. To reduce the workload, the clinicians adopt a sampling strategy and only a subset of all the lesion instances are marked in the dataset (see Fig~\ref{Fig2}). 

\begin{figure}
\centering
\includegraphics[width=1\columnwidth,height=\textwidth,keepaspectratio]{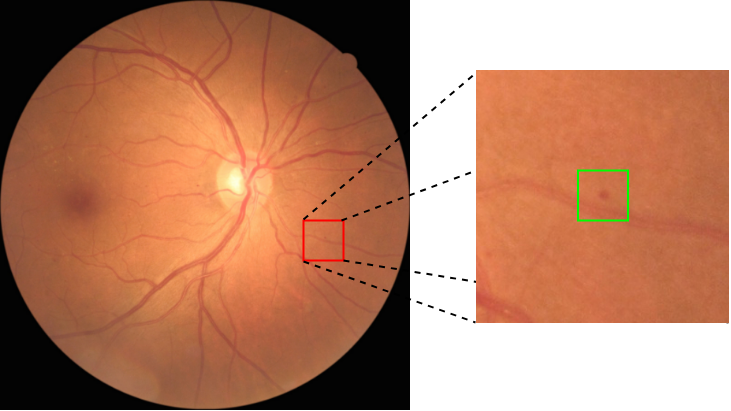}\caption{A mini lesion example: Microaneurysm with zoom-in view.}\label{fig1}
\end{figure}

Different from natural image datasets such as COCO\cite{lin2014microsoft}, VOC\cite{everingham2010pascal}, cityscapes\cite{cordts2016cityscapes} and Open Images\cite{kuznetsova2018open}, there are two major challenges for working with the DR dataset. 
First, the lesion samples may not be sufficient when ground-truth are only partially labeled and those real lesion instances without labels will not be used in the training process. To make it worse, unlabeled ground-truth will contribute to the negative sampling in the loss function, which may 
lead the model to move to the opposite direction.
Second, since the DR images are produced in high resolution format (e.g., 2136x3216) and the lesion instances have a very small footprint on each image, scaling down images to fit the input size (e.g., 800x800) of a normal CNN model will result in a significant quality loss, especially for those small lesions on the image. 

For the first challenge about the unlabeled ground-truth, we design a multi-round training algorithm based on the pseudo-labeling framework. We first train our model on the ground-truth with original manual labels to obtain a weak lesion detector in the first round, and then the detector will go through the training dataset to generate a lesion set detected by the current model. Note that the resulting lesion set now contains some lesions that have not been originally labeled.
We design a criterion to select some of the detected lesions as part of unlabeled ground-truth (UGT) set based on the confidence level of the detection. After that we employ the originally labeled ground-truth (LGT) together with newly found UGT 
to increase the number of training samples in the second round. We repeat the process of updating UGT and training new model, so as to improve the performance of the CNN detector.

To address the issue of small lesion size, we upgrade GPU during the training process so the CNN model can take an input size as large as the original image. Moreover, we construct a deeper feature pyramid network (FPN) with six scale layers to improve the expressivity of feature maps. With deeper and larger feature maps, the information of appearance and location for the small lesions will be abtracted more and thus the CNN models can produce better result. 
We investigate several detection methods based on CNN in this work and adopt Faster-RCNN and RetinaNet as the baselines. 
Experiment results show that our proposed multi-round algorithm can effectively discover those unlabeled ground-truth and significantly increase the training data. Furthermore, our model with large input size and deeper FPN considerably outperform the baselines.


\section{Related Work}
\subsection{Diabetic Retinopathy Diagnosis}
Wilkinson proposed International Clinical Diabetic Retinopathy and Diabetic Macular Edema Disease Severity Scales (ICDRDMEDSS) in 2003 as one of the international standards, in which there are 5 stages for diabetic retinopathy severity: 1) No DR, 2) Mild non-proliferative DR, 3) Moderate non-proliferative DR, 4) Severe non-proliferative DR, 5) Proliferative DR. 
The definition of each stage is indicated by location and the number of the following 10 lesion categories: 1) blot hemorrhages, 2) microaneurysms, 3) hard exudate, 4) cotton wool spot, 5) fibrous proliferation, 6) venous beading, 7) intraretinal microvascular abnormity (IRMA), 8) neovascularization, 9) vitreous hemorrhage, 10) venous loop \cite{wang2018diabetic}. Normally, mini lesions should be shown in detail on DR images with high-resolution format (see Fig~\ref{fig1}), especially for the small size lesion categories 1)-4), so that clinicians can have confidence in their diagnosis. 
\begin{figure}[ht]
\centering
\includegraphics[width=0.8\columnwidth,height=\textwidth,keepaspectratio]{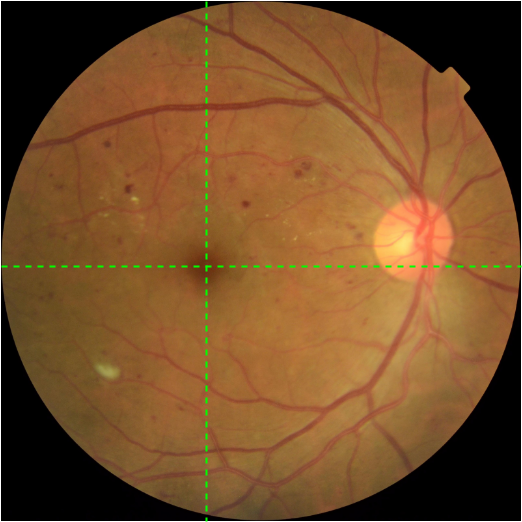}
\caption{DR image with 4 quadrants based on the center of marcula according to ICDRDMEDSS.}\label{fig5}
\end{figure}

The difference of the category, location and number of lesions on the fundus image signify different stages. For example, more than 20 hemorrhages in each of 4 quadrants and no signs of proliferative retinopathy indicates that the DB is in the 4th stage. Quadrant areas of fundus image are centered by macula (shown in Fig~\ref{fig5}). 
To examine the whole image and figure out all the numbers, categories and locations of these lesions is a labor intensive task for an ophthalmologist.
EyePACS \cite{cuadros2009eyepacs} is a well-known large scale dataset for DR Diagnosis in ICDRDMEDSS standard. It contains more than 100,000 fundus images and each image was labeled with an integer ranging from 0 to 4, indicating the stage of DR. In EyePACS, DR diagnosis is considered as a task of image classification. In fact, location and classification of lesions on fundus can show detail of DR diagnosis. Therefore lesion instance detection can provide an effective way to assist ophthalmologists to determine the condition of DR.

\subsection{CNN-based Object Detection}
CNN-based methods have become the main stream for visual object detection recently. A series of such methods with excellent performance have been proposed, such as Faster-RCNN \cite{ren2015faster}, SSD \cite{liu2016ssd}, RetinaNet \cite{lin2017focal} and YOLO \cite{redmon2016you} etc. 
Based on the number of stages in the detection process, these methods can be devided in two categories: one-stage method such as SSD, YOLO, RetinaNet, and two-stage methods such as Faster-RCNN. In one-stage detection, object localization and classification are produced at the same part of CNN. One of the disadvantages of the one-stage methods is the extreme unbalance between proposed positive targets and negative ones during the training process. But in RetinaNet, an one-stage detector, by using focal loss, lower loss is contributed by easy negative samples so that the loss is focusing on hard samples, which reduces the effect of unbalance on the loss value and thus improves the prediction accuracy. Unlike one stage methods, there are two branches in the structure of Faster-RCNN. One branch is the region proposal network (RPN), which can be viewed as a binary-label detector. RPN is designed to roughly identify the objectness part of one image. The binary-label detector will predict positive objectness results that will be used in the second step for the object label classfication and location regression. From previous studies, one-stage methods are faster and resource saving, and two-stage methods have better performance for various size object detection. In our work, we adopt Faster-RCNN and RetinaNet as the baselines in the experiments.

\subsection{Pseudo-Labeling}
Deep learning usually requires large amounts of labeled training data, but annotating data is costly and tedious. The framework of semi-supervised learning provides the means to use both labeled data and arbitrary amounts of unlabeled data for training. Recently, semi-supervised deep learning \cite{wu2017semi} has been intensively studied for standard CNN architectures. Pseudo-labeling is a semi-supervised learning method that can increase the performance of the CNN models by utilizing unlabeled ground-truth.
First proposed by Lee et. al. in 2013 \cite{lee2013pseudo}, the pseudo-labeling method uses a small set of labeled data along with a large amount of unlabeled data to improve the model’s performance. The technique of pseudo-labeling is simple and contains 3 basic steps. First, train the model on labeled data. Second use the trained model to predict labels on the unlabeled data, thus creating pseudo-labels. Third, combine the labeled data and the newly pseudo-labeled data in a new dataset that is used to train the data. Recently, pseudo-labeling has been used in various computer vision applications.
In particular, \cite{wu2017semi} and \cite{wang2016cost} use pseudo-labeling to enhance the model of image classification. Pathak et. al. \cite{pathak2017learning} applies automatically generated masks (pseudo-labels in the context) for image segmentation. 
Pseudo-labeling also achieves impressive results in the domain of visual object detection, such as \cite{liu2016fashion} and \cite{chen2018pseudo}. In our case, there are a certain number of missing labels in our DR lesion dataset and intuitively these instances could boost the performance of the lesion detector if they were used in the training process. To mine the missing ground-truth as much as possible, we propose an iterative training algorithm based on the basic pseudo-labeling framework. 

\subsection{Feature Pyramid Networks}
It has been proven that the construction of scale pyramids \cite{adelson1984pyramid} is an effective way to handle the fundamental challenge of recognizing objects at vastly various scales in computer vision. Feature Pyramid Network (FPN) was proposed in \cite{lin2017feature}. Just like image pyramid, Lin et al. build a scale pyramid structure upon CNN features as an enhancing component in recognition systems for visual objects of various scales. For both the one-stage methods and two-stage methods, detectors based on CNN with FPN achieve better result on large scale natural object detection dataset COCO \cite{lin2014microsoft}. The structure of standard FPN takes the last residual layer from the 4 stages of the backbone as input and then goes through a top-down pathway to construct 4 feature layers at different scales. The size ratio between adjacent layers is set to be 2 and the backbone is usually a ResNets \cite{he2016deep} without fully connected layers. In a standard FPN, the ratio of input to the largest feature scale is 4. Larger size feature can preserve more details of the objects, which is especially important for small instances. In original detection network in Faster-RCNN, a single-scale feature map is used and in this paper we adopt FPN in Faster-RCNN as one of the baselines. For RetinaNet, FPN is a basic part in the CNN model.

\section{Dataset and Method}
\subsection{Dataset}
Dataset collection is the first step for lesion detection methods based on CNN. We developed our own manual annotation tool base on VGG Image Annotator \cite{dutta2019vgg}, which can be easily applied to mark the bounding box and category of lesion instances on the DR images. 
The clinicians manually annotated lesions on the fundus images of the dataset. 
Our dataset contains $5,198$ images with a resolution of $2136\times3216$, including fundus pictures from 500 patients and covering all 5 severity stages. All the original images were preprocessed to remove the left and right hand side black parts with low pixel values. This helps the model focus on the fundus part with a new size of $2136\times2136$. There are 10 lesion labels as mentioned in the related work in our labeling tool and a flexible bounding box tool is provided to mark the location and category of each lesion. Each annotation box represents a single lesion instance and contains 5 values $(x,y,w,h,c)$, where $(x,y)$ are the coordinates of the upper-left corner of the ground-truth in a fundus image, $(w,h)$ is the dimension of the box and $c$ is the label of the lesion. The dataset was randomly divided into 4 equal parts and each part was handled by one clinician and then validated by 3 other clinicians. It is a labor-intensive task to mark all the lesions in the dataset. To reduce the workload, the clinicians label the lesion instances randomly with high confidence sampling strategy, which means that they only mark the significant lesions on each fundus image and some of the less obvious lesions are missed. 

\begin{table}[hb]
\centering
\begin{threeparttable}
\caption{Summary for the number of lesion categories 1-4.}\label{tab1}

\begin{tabular}{|c|p{1cm}<{\centering}|p{1cm}<{\centering}|p{1cm}<{\centering}|p{1cm}<{\centering}|}
\hline
\diagbox {Label\tnote{1}}{Set} & Total & Train & Validation \\
\hline
1& 18493 & 14720 & 3773 \\
\hline
2& 7703 & 6301 & 1402 \\
\hline
3& 9316 & 7403 & 1913 \\
\hline
4& 654 & 537 & 117 \\
\hline
\end{tabular}
\begin{tablenotes}
    \item[1] Labels correspond to the categories description order mentioned in the related work of diabetic retinopathy diagnosis.
  \end{tablenotes}
\end{threeparttable}
\end{table}

\begin{table}[hb]
\center
\caption{The average lesion-to-image ratios for categories 1-4.}\label{tab2}
\begin{tabular}{|c|c|c|c|c|}
\hline
Label&1&2&3&4 \\
\hline
Ratio&0.07244\%&0.05390\%&0.31672\%&0.23976\%\\
\hline
\end{tabular}
\end{table}

Table~\ref{tab1} summarizes the number of instances for lesion categories 1-4. 
The remaining categories 5-10 only have 34, 15, 25, 49, 14 and 1 instances respectively, constituting a tiny fraction ($0.3\%$) of the dataset.
The first four categories are more valuable for early diagnosis of DR because they are indicators for the first three stages of severity \cite{wang2018diabetic}. Compared with the other six categories of lesions, labeling of categories 1-4 are more challenging and
these lesions tend to be very small on a fundus image. The average lesion-to-image ratios for categories 1-4 are listed in Table~\ref{tab2}. 


\subsection{Pesudo-labeling with Iteration Training Process}
The basic idea of pseudo-labeling is to train a CNN model that can automatically find and annotate the unlabeled ground-truth and add the result to the training set to retrain the CNN model.
Frist, we design a two-round training process to implement this idea. The first round is to train a weak lesion detector on the dataset with the original manually labeled ground-truth. The weak lesion detector will then go through the training set 
and a certain amount of unlabeled ground-truth will be discovered. 
\begin{figure}
\centering
\includegraphics[width=0.7\columnwidth,height=\textwidth,keepaspectratio]{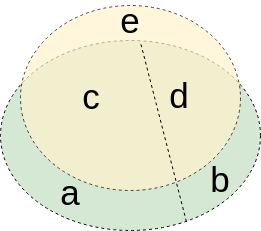}
\caption{Venn diagram for the relationship between label sets: $a,b,c,d,e$ are seperated by dotted line, $a$ is the undetected lesion instances set with labels, $b$ is the undetected lesion instances set without labels, $c$ is the detected lesion instances set with labels, $d$ is the detected lesion instances set without labels and $e$ is the false positive set. Union $a$$\cup$$b$$\cup$$c$$\cup$$d$ is ground-truth set $G$, $a$$\cup$$c$ is LGT set $D$, $b$$\cup$$d$ is the UGT set $R$, $c$$\cup$$d$$\cup$$e$ is $W$. We aim to get a set $X$ that approximates $d$.}\label{fig5_VD}
\end{figure}
In our work, we refer to the lesion instances with manual labels as labeled ground-truth (LGT) and those without manual labels as unlabeled ground-truth (UGT).
During the second round, the CNN model will be trained on both LGT and UGT.
Not all the detection result from the first round can be added into the UGT set as some of the newly discovered lesions are just false positive. We need a criterion to determine
whether a newly detected lesion instance should be included in UGT set. 
We define LGT set as $D=\{g_1, ..., g_i\}$ and UGT set as $R=\{g_{i+1}, ..., g_j\}$. The intersection between $D$ and $R$ is empty, i.e., $D$$\cap$$R=\emptyset$. The collection of all the ground-truth is $G=D$$\cup$$R$. 
The lesions found by weak detector on the training set can be defined as $W=\{w_1, ..., w_k\}$. The relationship among $(G,D,R,W)$ is shown in a Venn diagram (Fig~\ref{fig5_VD}). We can see that $W$ contains part of $G$ and false positive set (region $e$ shown in Fig~\ref{fig5_VD}). We aim to get a set $X$ similar to $d$ (unlabeled true lesions set) in the Venn diagram after the first round of training. Another issue is that we need to come up with a criterion to identify those real unlabeled lesions from $W$. 
In fact, there is confidence value for each label of ground-truth. For $D$, the confidence of $g_m$ is $s^{g}_{m}=1$, $1$$\leq$$m$$\leq$$i$. When the weak detector produces $W$, it will generate a probability $s^{w}_{n}\in{(0,1)}$ for each instance $w_n$ at the same time, $1$$\leq$$n$$\leq$$k$, $0<s<1$. Intuitively, $w_n$ is more likely to be a ground-truth as the confidence level $s^{w}_{n}$ increases. We define a criterion to select UGT in $W$:
\begin{equation}
C_n^w=
\begin{cases}
1,&\text{if ($s^{w}_{n}$$>$$P$) and ($IoU_{n}^{g_{(1, ..., i)}}$$<$$0.05$)}\\
0,&\text{otherwise,}
\end{cases}
\label{e1}
\end{equation}
where $C_n^w=1$ represents that $w_n$ in $W$ is ground-truth, $s^{w}_{n}$ is the predicted probability for $w_n$. $P$ is the threshold for the instance to be a true positive. $IoU_{n}^{g_(1, ..., i)}$ indicates the intersection-over-union (IoU) between $w_n$ and $g_m$, $g_m$$\in$$D$. 0.05 indicates that the UGT should have low IoU with LGT. 
In the second round, $D$ and $X$ are combined as the ground-truth set to train the CNN model. 
\begin{figure*}[ht]
\centering
\includegraphics[width=1.8\columnwidth,height=\textwidth,keepaspectratio]{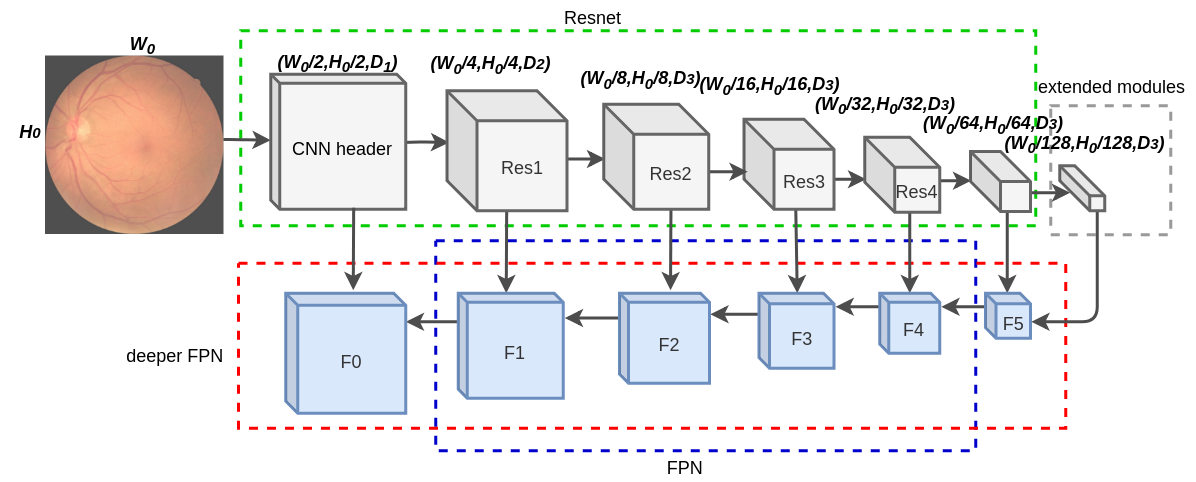}
\caption{The CNN architecture with large input size and deeper FPN. The backbone is Resnet and there are six feature maps with different scales in FPN.}\label{fig2}
\end{figure*}
\begin{algorithm}[h]  
    \caption{Iterative Algorithm for the Multi-round Training}
    \label{algorithm1}  
    \begin{algorithmic}[1]
        \REQUIRE  
            $f(D)$: CNN detector;  
            $D$: LGT set;
            $X$: UGT set;
        \ENSURE $D_*$: union set of LGT and UGT;
        \STATE 
        initialize UGT set $X=\emptyset$;
        $D_*=D$;
        \REPEAT
        \STATE merge $X$ into $D_*$;
        \STATE train model $f(D)$ on the union set $D_*$;
        \STATE compute $f(D)$ on training set and get a result set $T$;  
        \STATE obtain a subset $X$ from $T$ according to the criterion in equation~(\ref{e1});
        \UNTIL{$L(X)\leq M$}
    \end{algorithmic}
\end{algorithm}

After the second-round training, we notice that the retrained CNN model can still detect more unlabeled ground-truth, suggesting that the model can benefit from more rounds of training. 
To this end, we design an iterative algorithm 
for the multi-round training to make the set $X$ approaching $d$ as much as possible, as shown in Fig~\ref{fig5_VD}. In algorithm~\ref{algorithm1}, to make the iteration converge, we increase the value of threshold $P$ as the number of iteration increases. $L(X)$ represents the number of labels in $X$ and $M$ is minimum number of labels in $X$.


\subsection{Large Input Size and Deeper FPN}\label{AA}
Fig~\ref{fig2} illustrates the backbone architecture of our model with larger input size and deeper FPN. The backbone of baselines is a normal Resnet and images with the short side length more than 800 pixels should be zoomed out to fit the input size. 
In our case, the resolution of DR images is 2136x2136 and they should be rescaled to almost one quarter of the original size before fed into the CNN model. Image scaling can be interpreted as a form of image resampling or image reconstruction from the view of the Nyquist sampling theorem \cite{vaidyanathan2001generalizations}. According to the theorem, down-sampling to a smaller image from a higher-resolution original can only be carried out after applying a suitable 2D anti-aliasing filter to prevent aliasing artifacts. Decreasing the pixel number (scaling down in our case) usually results in a visible quality loss, especially for mini lesions on the DR images, the detail of the feature could be lost after the operation of scaling down. We set the input size of the backbone as large as the size of original DR image. The Resnet \cite{targ2016resnet} is employed as the backbone and the anchor scales are shrinked to fit the size of mini lesions.

A normal Resnet backbone contains a head module $H_{0}$ and 4 Resnet modules \{$R_{0}$, $R_{1}$, $R_{2}$, $R_{3}$\}. 
In this paper, we put the head module in the process of FPN construction, which will produce larger feature map ($F_0$ in Fig~\ref{fig2}) for small object detection. Moreover, as Resnet module is more effective for extraction of object appearance and location than single CNN layer alone, we extend two more Resnet modules \{$R_{4}$, $R_{5}$\} to construct a deeper FPN.

\section{Experiments}

In this section, we will first introduce the evaluation metrics for our experiments and then describe the details of hyperparameters and analysis of the results. We select all images with small lesion categories $1-4$ as the experimental dataset. During the experiments, we randomly divide the data into two sets, one for training while the other for validation with a 4:1 ratio. The number of lesion instances of both sets are shown in the last two columns of Table~\ref{tab1}. 
Due to the incompleteness of ground-truth labels in the dataset, both the validation set and training set contain a certain amount of unlabeled ground-truth. In our experiments, we aim to automatically detect the ground-truth and label them as much as possible from the validation set. Thus it is more reasonable to use the sensitivity as the quality metric on the validation set. In our experiments, horizontal flipping is applied during the training for data augmentation. We adopt Faster-RCNN with FPN and RetinaNet in  MMdetection \cite{mmdetection} based on pytorch as the baselines and the experiment is performed on 2 NVIDIA Titan RTX GPUs and the memory size in each GPU is 24G.

\subsection{Evaluation Metrics and Parameter Setting}
Standard evaluation metrics for natural object detection is based on IoU ratio only \cite{ren2015faster}. The IoU ratio between a true positive object and the ground-truth should be above the threshold 0.5. 
In our DR dataset, the bounding boxes of ground-truth lesion instances usually contain a large surrounding area indicated by the green rectangle shown in Fig~\ref{fig3} while 
the main body of a lesion (indicated by the blue rectangle) is usually much smaller and located in the center region of the bounding box. A predicted object that has an IoU ratio less than 0.5 but contains the center region (main body of the lesion) is still acceptable in our application. 
We present a center-focus (CF) target criterion \cite{chen2019mini} to define positive instances, in which a proposed instance is considered positive when the IoU ratio is more than 0.1 and the proposed rectangle contains the center point of the ground-truth. 
This criterion is used in both training and validation. 
\begin{figure}[hb]
\centering
\includegraphics[width=0.8\columnwidth,height=\textwidth,keepaspectratio]{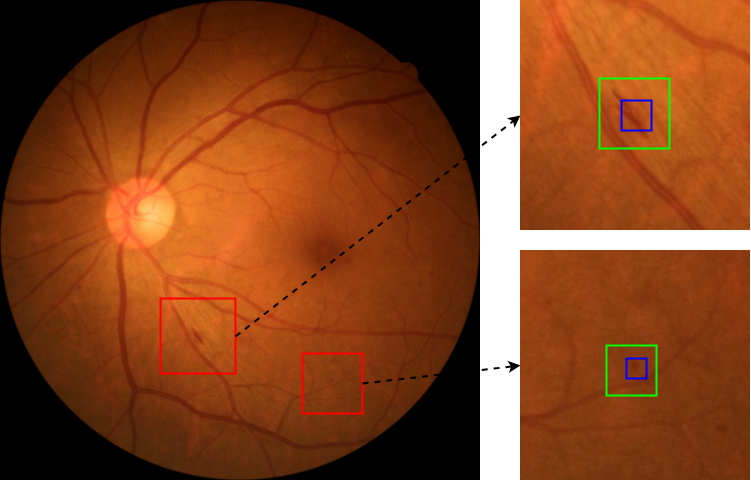}
\caption{Red rectangle parts of the left image are zoomed in and shown on the right. Green rectangles are the ground-truth boxes and blue ones are proposal targets.}\label{fig3}
\end{figure}

Resnet101 is employed as the backbone, which has been pretrained on Imagenet \cite{deng2009imagenet}. We set the total training epoches as twelve. The base anchor number of each location in one layer is $12$. In our work, we use a small anchor scale list $\{1, 2, 4, 8\}$ and an anchor ratio list $\{0.5, 1, 2\}$ for all the three methods to fit the size of small lesions. During the validation step, we set the max prediction number of one image to 100 and the confidence score threshold is 0.1. The same parameter settings are used for all the following experiments.
\subsection{Large Input Size and Deeper FPN}


First, we conduct a series of experiments to study the effect of input size on the performance of the model. In the experiments, the input size varies from 800 (pixels) to 2,000 with a step size of 200. 
From the results in Table~\ref{tab_size}, we can see that large input size can improve the performance at a considerable rate.
In the experiments to verify the effectiveness of deeper FPN, the input image is reconstructed with a resolution of $2136\times2136$.  In a normal FPN, there are four layers for region proposal and region of interest (RoI) classification. In the structure of deeper FPN, 
the number of feature pyramid scales is extended to six and the largest feature map $F_0$ can be used for small lesion detection. The results of sensitivity in CF criterion are shown in Table~\ref{tab3}. For lesion categories 1-4, the performance of our proposed deeper FPN are superior to the baselines for both one-stage method RetinaNet and two-stage method Faster-RCNN. The results shows that the large feature map is more suitable for smaller lesion detection, especially for lesions of categories 1-2 with smaller area ratios.

\begin{table}
\begin{threeparttable}
\caption{Performance of lesion detectors at different input sizes. }\label{tab_size}
\centering
\begin{tabular}{|c|p{0.9cm}<{\centering}|p{0.9cm}<{\centering}|p{0.9cm}<{\centering}|p{0.9cm}<{\centering}|p{0.8cm}<{\centering}|}
\hline
\diagbox{Size\tnote{1}}{Label} & 1& 2& 3 & 4 &Model\tnote{2}\\
\hline

\multirow{2}*{800}&80.57\%&80.08\%& 83.36\% & 67.39\%&F\\
\cline{2-6}
                 ~&79.93\% & 77.65\% & 84.05\% & 68.55\%&R \\
\hline

\multirow{2}*{1000}&80.98\%&80.21\%& 84.43\% & 68.63\%&F\\
\cline{2-6}
                 ~& 80.22\% & 79.89\% & 84.34\% & 69.01\%&R \\
\hline

\multirow{2}*{1200}&81.75\%&81.71\%& 84.66\% & 68.89\%&F\\
\cline{2-6}
                 ~& 80.98\% & 82.44\% & 84.32\% & 69.23\%&R \\
\hline

\multirow{2}*{1400}&83.95\%&83.21\%& 85.34\% & 70.32\%&F\\
\cline{2-6}
                 ~& 83.12\% & 84.11\% & 86.33\% & 71.44\%&R \\
\hline

\multirow{2}*{1600}&85.12\%&85.43\%& 87.15\% & 72.54\%&F\\
\cline{2-6}
                 ~& 84.65\% & 85.32\% & 88.31\% & 73.63\%&R \\
\hline

\multirow{2}*{1800}&86.62\%&86.13\%& 88.25\% & 73.10\%&F\\
\cline{2-6}
                 ~& 84.96\% & 86.14\% & 89.41\% & 74.32\%&R \\
\hline

\multirow{2}*{2000}&86.83\%&86.33\%& 88.51\% & 73.13\%&F\\
\cline{2-6}
                  ~& 85.21\% & 85.35\% & 89.79\% & 74.60\% &R\\
\hline

\end{tabular}
\begin{tablenotes}
    \item[1] The value of size represents the shorter side length of input image.
    \item[2] F is the abbreviation of Faster-RCNN with FPN and R is for RetinaNet.
  \end{tablenotes}
\end{threeparttable}
\end{table}

\begin{table}
\centering
\begin{threeparttable}
\caption{Performance of detectors at different training conditions. The input size is 2136*2036.}\label{tab3}

\begin{tabular}{|c|p{1cm}<{\centering}|p{1cm}<{\centering}|p{1cm}<{\centering}|p{1cm}<{\centering}|}
\hline
\diagbox{Model}{Label} & 1 & 2 & 3 & 4 \\
\hline
F+FPN& 86.85\% & 86.33\% & 88.60\% & 73.13\% \\
\hline
RetinaNet& 85.43\% & 85.23\% & 89.97\% & 74.60\%\\
\hline
\textbf{F+DFPN\tnote{1}}&89.83\%&90.43\%& 89.51\% & 74.10\%\\
\hline
\textbf{R+DFPN}&88.65\%&87.49\%& 90.01\% & 74.93\%\\
\hline
\textbf{F+DFPN+MR\tnote{2}}&95.83\%&93.11\%& 95.26\% & 87.45\%\\
\hline
\textbf{R+DFPN+MR}&94.98\%&92.51\%& 96.51\% & 88.32\%\\
\hline
\end{tabular}
\begin{tablenotes}
    \item[1] DFPN means deeper FPN.
    \item[2] MR is multi-round training.
  \end{tablenotes}
\end{threeparttable}
\end{table}

\begin{table}
\caption{Number of UGT with different threshold $P$ in the second round training.}\label{tab4}
\centering
\begin{tabular}{|c|p{0.8cm}<{\centering}|p{0.8cm}<{\centering}|p{0.8cm}<{\centering}|p{0.8cm}<{\centering}|p{0.8cm}<{\centering}|}
\hline
\diagbox{$P$}{Label} & 1 & 2 & 3 & 4 &Model\\
\hline
\multirow{2}*{0}& 1821 & 863 & 842 & 631 &F \\
\cline{2-6}
~& 1638 & 785 & 930 & 682 &R \\
\hline

\multirow{2}*{0.1}& 1400 & 681 & 757 & 499&F \\
\cline{2-6}
~& 1529 & 601 & 689 & 554 &R\\
\hline

\multirow{2}*{0.2}&  1014 &  523 & 588 & 364 &F\\
\cline{2-6}
~&  932 &  478 & 643 & 388 &R\\
\hline

\multirow{2}*{0.3}& 805 & 431 & 490 & 303 &F\\
\cline{2-6}
~& 744 & 405 & 523 & 309 &R\\
\hline

\multirow{2}*{0.4}& 670 & 364 & 417& 258 &F\\
\cline{2-6}
~& 598 & 309 & 433& 276 &R\\
\hline

\multirow{2}*{0.5}& 569 &  307 & 357& 215 &F\\
\cline{2-6}
~& 531 &  276 & 381& 229 &R\\
\hline

\multirow{2}*{0.6}&489 &  252 & 305 & 186 &F\\
\cline{2-6}
~&409 &  221 & 325 & 165 &R\\
\hline

\multirow{2}*{0.7}& 420 &  196 & 232 & 146 &F\\
\cline{2-6}
~& 362 &  163 & 253 & 153 &R\\
\hline

\multirow{2}*{0.8}& 347 &  135 & 195 & 123 &F\\
\cline{2-6}
~& 287 &  135 & 195 & 136 &R\\
\hline

\multirow{2}*{0.9}& 238 &  87 & 122 & 93 &F\\
\cline{2-6}
~& 199 &  61 & 148 & 105 &R\\
\hline

\end{tabular}
\end{table}

\begin{table}
\caption{Summary of sensitivities on 4 categories at different level of confidence threshold.}\label{tab5}
\centering
\begin{tabular}{|c|p{0.8cm}<{\centering}|p{0.8cm}<{\centering}|p{0.8cm}<{\centering}|p{0.8cm}<{\centering}|p{0.8cm}<{\centering}|}
\hline
\diagbox{$P$}{Label} & 1 & 2 & 3 & 4 &Model\\
\hline
\multirow{2}*{0}& 93.98\% & 90.45\% & 95.11\% & 86.12\%&F \\
\cline{2-6}
               ~& 93.54\% & 89.91\% & 95.32\% & 86.44\%&R \\
\hline
\multirow{2}*{0.1}& 94.14\% & 90.51\% & 94.77\% & 86.06\%&F \\
\cline{2-6}
                 ~& 93.89\% & 90.87\% & 95.03\% & 86.14\% &R\\
\hline
\multirow{2}*{0.2}&  94.01\% &  91.43\% & 94.99\% & 85.85\% &F\\
\cline{2-6}
                 ~&  93.87\% &  \textbf{91.55\%} & \textbf{95.91\%} & 86.43\% &R\\
\hline
\multirow{2}*{0.3}& \textbf{94.92\%} & \textbf{92.46\%} & \textbf{95.65\%} & \textbf{86.17\%} &F\\
\cline{2-6}
                 ~& \textbf{94.54\%} & 91.53\% & 95.88\% & \textbf{87.12\%} &R\\
\hline
\multirow{2}*{0.4}& 93.41\% & 90.48\% & 94.61\%& 84.71\% &F\\
\cline{2-6}
                  ~& 92.91\% & 90.12\% & 94.98\%& 85.35\% &R\\
\hline
\multirow{2}*{0.5}& 92.89\% &  90.30\% & 93.89\%& 82.33\% &F\\
\cline{2-6}
                 ~& 92.45\% &  90.12\% & 94.41\%& 83.46\% &R\\
\hline
\multirow{2}*{0.6}&92.03\% &  90.55\% & 93.55\% & 81.03\% &F\\
\cline{2-6}
                 ~&91.78\% &  90.45\% & 93.98\% & 82.11\% &R\\
\hline
\multirow{2}*{0.7}& 91.51\% &  90.11\% & 91.89\% & 80.60\% &F\\
\cline{2-6}
                ~& 91.31\% &  90.51\% & 92.65\% & 81.64\% &R\\
\hline
\multirow{2}*{0.8}& 91.02\% &  90.34\% & 92.01\% & 78.31\% &F\\
\cline{2-6}
               ~& 90.45\% &  88.34\% & 92.33\% & 79.40\% &R\\
\hline
\multirow{2}*{0.9}& 90.57\% &  90.32\% & 90.13\% & 76.32\% &F\\
\cline{2-6}
                 ~& 90.14\% &  87.34\% & 91.43\% & 77.98\% &R\\
\hline
\end{tabular}
\end{table}

\subsection{Multi-round Training}
As the ablation experiments in previous section have verified the effectiveness of large input size and deeper FPN to detect small lesions on DR images, we implement the multi-round training on the structure of large Faster-RCNN with deeper FPN. After the first-round training, the weak lesion detector will produce the newly discovered lesion set $W$. In the criterion defined in equation~(\ref{e1}), the threshold $P$ of the confidence score to classify a candidate $w$ as a ground-truth is a variable. Different values of $P$ will result in the set $X$ of different size. Table~\ref{tab4} 
shows the ratio of the cardinality of UGT to $X$ for different values of $P$ with a step size of 0.1. We observe that lower values of $P$ add more lesion instances to UGT set.

In the following rounds, $X$ is combined with $D$ as the ground-truth to be used in the training process. 
Specially we study the results at different values of threshold $P$ for the second round training. 
Table~\ref{tab5} shows the sensitivity results on various $P$ value. 
The performance is better for lower values of $P$ as more lesion instances are added to the training and validation set 
in the following round training. 
The model can learn more features from the UGT set, especially for the categories with less manual annotations such as Cotton-wool Spot. But there is a limit for the possible benefit from UGT.
From Table~\ref{tab5}, we can see that a larger UGT set with low confidences, i.e, $P=0.1$ or $P=0.0$ will not result in better sensitivity, because some of these UGT could be false positives. In the experiments of multi-round training, we set $P=0.3$ and $M=100$ in Algorithm~\ref{algorithm1}. In the training process of Faster-RCNN, the number of max iteration reaches 3 and for RetinaNet, the number reaches 4. 
The final precision results of multi-round training are listed in the last two row of Table~\ref{tab3} and shows that the iterative algorithm can improve the performance with pseudo-labeling. 
\begin{table}
\caption{The number of UGT in the final round training.}\label{tab6}
\centering
\begin{tabular}{|c|p{1cm}<{\centering}|p{1cm}<{\centering}|p{1cm}<{\centering}|p{1cm}<{\centering}|}
\hline
\diagbox{Model}{Label}& 1 & 2 & 3 & 4 \\
\hline
F& 1043& 632 & 612 & 417 \\
\hline
R& 1136 & 693 & 663 & 459\\
\hline
\end{tabular}
\end{table}
\section{Conclusion}
In this paper we introduce an iterative training algorithm for the semi-supervised method of pseudo-labeling to resolve the issue of incomplete manual labels in the dataset. The proposed solution leads to significant improvement over the baselines. 
Moreover, we study the effect of input size of CNN model and propose a deeper FPN structure to detect small lesion instances on DR images.
Extensive experiments have demonstrated the superiority of our proposed ideas. The large CNN feature maps can preserve the details of small lesions 
and thus are more effective for object detection in both one-stage and two-stage methods. Moreover, multi-round strategy can reduce the dependence on the manual annotation when training CNN model.


\bibliographystyle{./bibliography/IEEEtran}
\bibliography{ref}

\end{document}